\documentclass{article}

\usepackage[preprint]{neurips_2021}
\usepackage{neurips_2021}

\usepackage[utf8]{inputenc} 
\usepackage[T1]{fontenc}    
\usepackage{hyperref}       
\usepackage{url}            
\usepackage{booktabs}       
\usepackage{amsfonts}       
\usepackage{nicefrac}       
\usepackage{microtype}      
\usepackage{xcolor}         

\title{Prediction of Household-level Heat-Consumption using PSO enhanced SVR Model}

\author{%
  Satyaki Chatterjee$^1$, Siming Bayer$^2$, and Andreas Maier$^3$
\\
  Pattern Recognition Lab\\
  Friedrich-Alexander-Universität Erlangen-Nürnberg\\
  Martensstr. 3, 91058 Erlangen, Germany\\
  \texttt{$^1$satyaki.chatterjee@fau.de, $^2$siming.bayer@fau.de, $^3$andreas.maier@fau.de} \\
}

\begin{document}

\maketitle

\begin{abstract}
  
  In combating climate change, an effective demand-based energy supply operation of the district energy system (DES) for heating or cooling is indispensable. As a consequence, an accurate forecast of heat consumption on the consumer side poses an important first step towards an optimal energy supply. However, due to the non-linearity and non-stationarity of heat consumption data, the prediction of the thermal energy demand of DES remains challenging. 
  In this work, we propose a forecasting framework for thermal energy consumption within a district heating system (DHS) based on kernel Support Vector Regression (kSVR) using real-world smart meter data. 
  Particle Swarm Optimization (PSO) is employed to find the optimal hyper-parameter for the kSVR model which leads to the superiority of the proposed methods when compared to a state-of-the-art ARIMA model. The average MAPE is reduced to 2.07\% and 2.64\% for the individual meter-specific forecasting and for forecasting of societal consumption, respectively.
\end{abstract}

\section{Introduction}
According to the United Nation Environment Program, `cities are a key contributor to climate change', as urban areas are `responsible for 75\% of global CO2 emissions' \cite{UNEP_1}. One of the effective weapons in combating climate change is the district energy system (DES) \cite{UNEP_2}, which implies the need for demand-driven thermal energy supply based on real consumption. However, utilities tend to over-supply the DES network to date, as the security of the energy supply must be ensured and the accumulated energy consumption acquired by the consumer side installation of smart meters is solely used for billing purposes. Hence, the behavior of the end-consumer remains largely unknown for the utilities while operating the DES network. Consequently, an accurate forecast of thermal energy demand poses the logical first step towards an optimized energy supply of the entire network. The focus of our work is the energy demand forecast of district heating systems (DHS), as it is the majority of the installed DES networks worldwide. 

Although the performance of statistical forecast methods, e.g.~the AR\cite{ar}, ARIMA\cite{bib4,arima1} or SARIMA \cite{bib10} models already demonstrate promising results in general, the prediction accuracy of energy load is limited due to the nonlinearity in the underlying time series. Recently, a model using the artificial neural network (ANN), e.g.~\cite{bib5}, is proposed to address the nonlinearity present heat consumption/load data. However, the training of ANN is often challenging, as it has the tendency of being trapped in local optima \cite{bib2}. Deeper models like Long Short Time Memory (LSTM) has the capability to model the underlining trend, seasonality, residuals, and external factors \cite{lstm}, however the training of such model prerequisites a large amount of data that can not be provided by the vast majority of the utilities. Traditional machine learning methods such as Support Vector Regression (SVR) \cite{bib11}, which is equipped with the inherent capability to deal with nonlinearity \cite{bib7}, present a compromise. SVR-based methods demonstrate their superiority to other machine learning models for heat load prediction in \cite{saguna, gao, svrcal}. As the performance and the usability of SVR are greatly affected by the choice of the model hyper-parameters, evolutionary algorithms such as Particle Swarm Optimization (PSO) \cite{bib12} are employed to determine the optimal hyper-parameters automatically. Previously, PSO-SVR is proposed in \cite{bib2} for load prediction of a DHS based on the daily heat load data during the heating season acquired from the production plant that demonstrates promising results. 

In this paper, the main contributions of our work are: (1) a machine learning-based forecast framework enables the use of accumulated heat consumption data acquired with smart meters that are installed on the end-consumer side; (2) the formulation of a PSO-kSVR model that has the inherent capability to deal with the nonlinearity of the underlying data; and (3) to the best of our knowledge, exploring the use of PSO-SVR and accumulated heat consumption for the forecast of energy load with real-world smart meter data on the house-hold level for the first time.

\section{Accumulated Consumption Data and Data Analysis}
\label{sec:heat_data}
The smart meter data used in this work is provided by a danish utility with an hourly temporal resolution. It is a small size municipality with 3808 consumers. As the smart meters are installed one after the other instead of all at once, the time period of the raw data spans from 6 to 12 months. Additionally, public available meteorological data of the same municipality at the same time interval are utilized to incorporate the weather conditions.
\begin{figure}[h!]
\subfigure[]{\includegraphics[width=7cm]{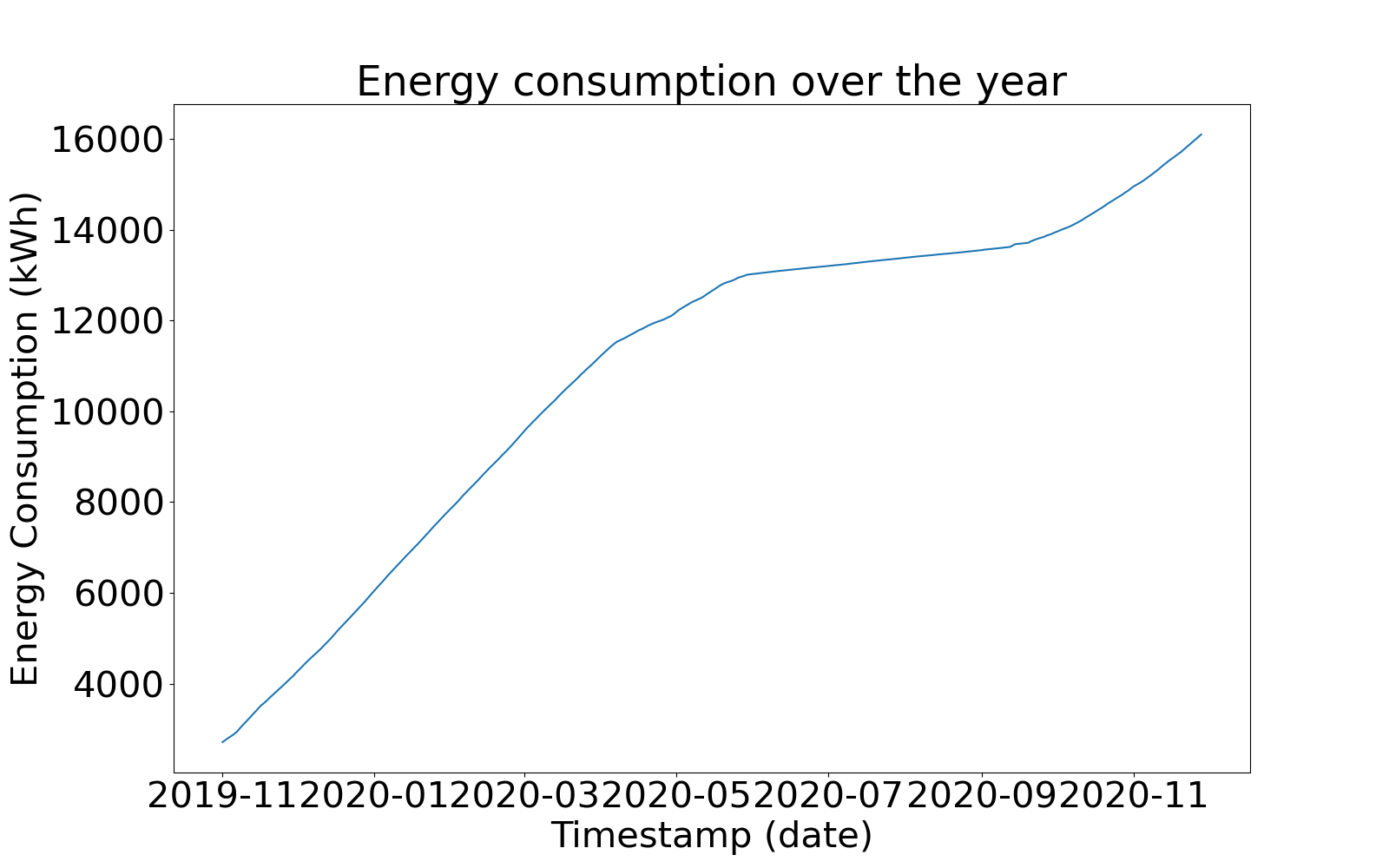} \label{fig:en_year}} 
\subfigure[]{\includegraphics[width=7cm]{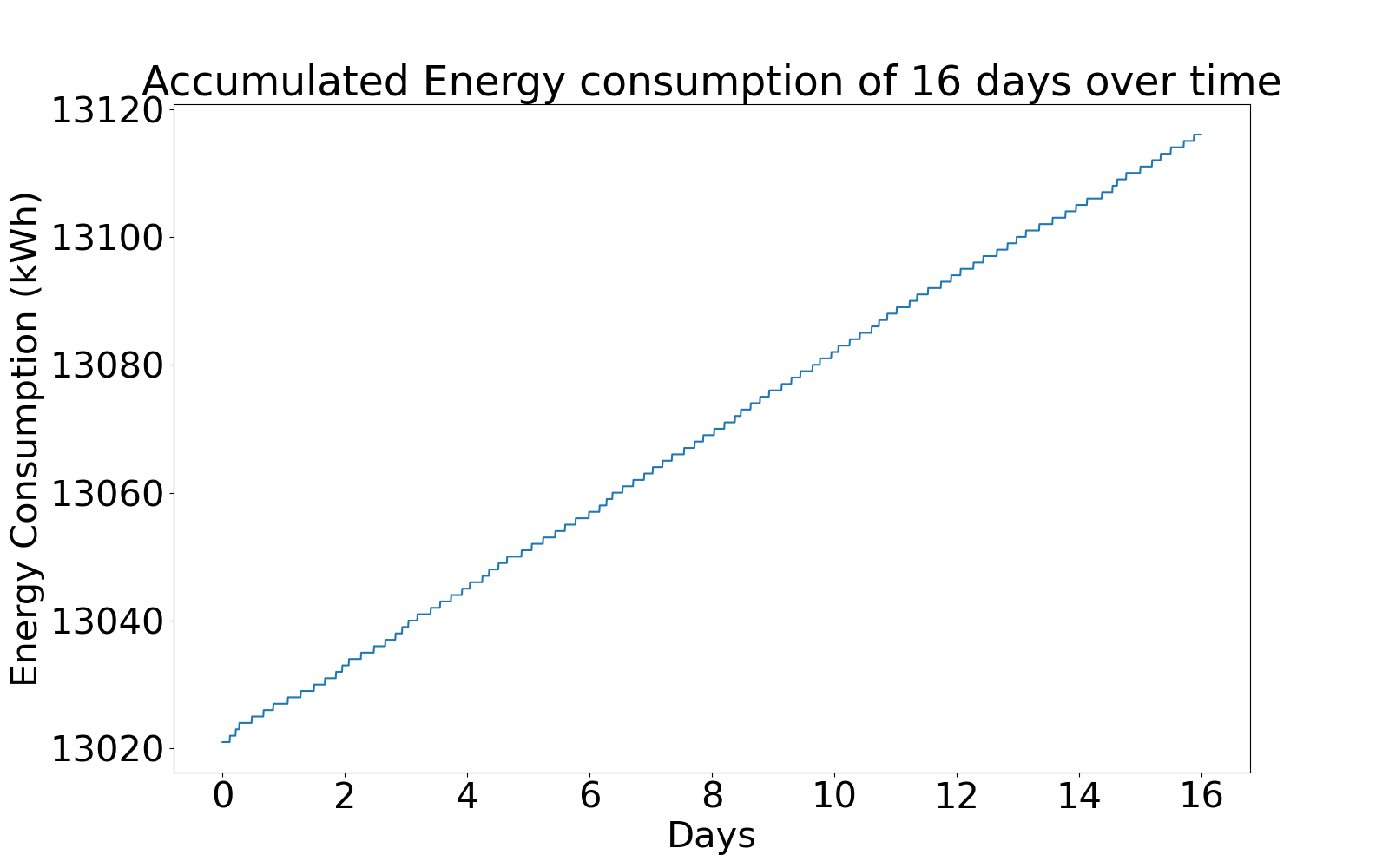}\label{fig:en_16}} 
\caption{(a) Pattern of accumulated heat consumption for a year; (b) Pattern of accumulated heat consumption for 16 days}
\end{figure}
Stationarity tests and decomposition methods are applied to analyse the data property. Basically, the accumulated energy consumption over a year is non-linear and non-stationary (refer to Fig.~\ref{fig:en_year}). However, it shows a linear trend and stationarity in a 16 days window (refer to Fig.~\ref{fig:en_16}). Furthermore, the accumulated consumption data of 16 days can be decomposed into a linear trend, daily seasonality, and residuals. From a signal processing point of view, the trend can be considered as the low-frequency component, whereas the seasonality and residuals pose the high-frequency component of the data.

Additionally, we performed a correlation analysis to identify the most descriptive features based on historical consumption and the feel-like temperature. In comparison to average temperature, feel-like temperature represents the maximum and minimum temperature, as well as wind speed and humidity, therefore has a higher descriptive character. 
In order to identify the appropriate lag length of the historical consumption, PARCOR-coefficients are computed and analysed by auto-correlating the daily data. The lag length of the corresponding feel-like temperature is identified by performing cross-correlative analysis with the accumulated consumption data. The aforementioned analysis on a set of randomly selected smart meters data demonstrates the insignificance of lag features beyond a lag of one hour. Thus, historical consumption and feel-like temperature with the lag of 1 hour, denoted as $h^{t-1}$ and $\theta^{t-1}$ are identified as the most descriptive features for our prediction task. 

\section{PSO-SVR framework}

The input data for our PSO-kSVR framework are time series of accumulated heat consumption acquired by smart meters (refer to Fig.~\ref{fig:en_year}) and is split into training, validation, and test data. The intuition behind this choice is that the accumulation essentially acts as integration. From the signal processing perspective, integration can be considered as a low-pass filter. Therefore, the high-frequency noise components in the accumulated consumption data are suppressed and thus the enhanced signal-to-noise ratio leads to enhanced performance in prediction \cite{HT}. As the raw data of the smart meter are usually unevenly spaced time series due to the transmission errors within wireless communication networks, the input data are first interpolated and resampled to evenly spaced time series. Subsequently, a set of descriptive features, that are identified by employing data analysis techniques (detailed in Sec.~\ref{sec:heat_data}), are extracted from the preprocessed data.
In order to model the seasonality and residuals of the underlying heat consumption data and find the non-linear correlation between weather and heat consumption, kernel-SVR (kSVR) with Radial Basis Function (RBF) kernel is employed. Hereby, the heat load prediction is formulated as an optimization task, which minimizes the square of the L2-norm of the coefficient vector ($\vec{w}$) of the features such that the error between the target and predicted accumulated consumption is within a margin of error.

As the choice of the hyper-parameters $\epsilon$, $C$, and $\gamma$ affect the model performance greatly, PSO is employed to estimate the optimal hyper-parameters for kSVR. The validation data are used to update the configuration of the particles in a three-dimensional search space. The final predicted consumption for each discrete time step is estimated as a time shift corrected the first-order derivative of the model output, i.e.~predicted accumulated energy. 

\section{Experiments and Result}\label{sec5}
Two types of experiments are conducted in this work to evaluate the accuracy and robustness of the proposed PSO-kSVR prediction framework:
\begin{description}
\item [Typ I:] qualitative and quantitative evaluation with a different subset of each meter-specific dataset aiming at analyzing the performance and applicability of our method for different smart meters, henceforth \textbf{meter-specific evaluation}. 
\item [Typ II:] comparative experiments aiming at comparing our proposed method with the state-of-the-art ARIMA model and analyzing the effect of different seasons/months. For this purpose, all meters that have 12 months of historical data are summed up to reflect the societal consumption of the municipality. Therefore, this comparative study is referred to as \textbf{societal evaluation} henceforth.
\end{description}

For the both settings, a similar configuration of the PSO algorithm that is introduced in \cite{bib1} is utilized, facilitating a prediction for a window of 24 hours. The evaluation metrics for the quantitative results are root mean squared error (RMSE) and mean absolute percentage error (MAPE). The input data are divided into subsets containing 16 days of consumption and weather data. The split ratio of training, validation, and test data is 14:1:1.

\begin{figure}[h!]
    \includegraphics[width=14.2cm]{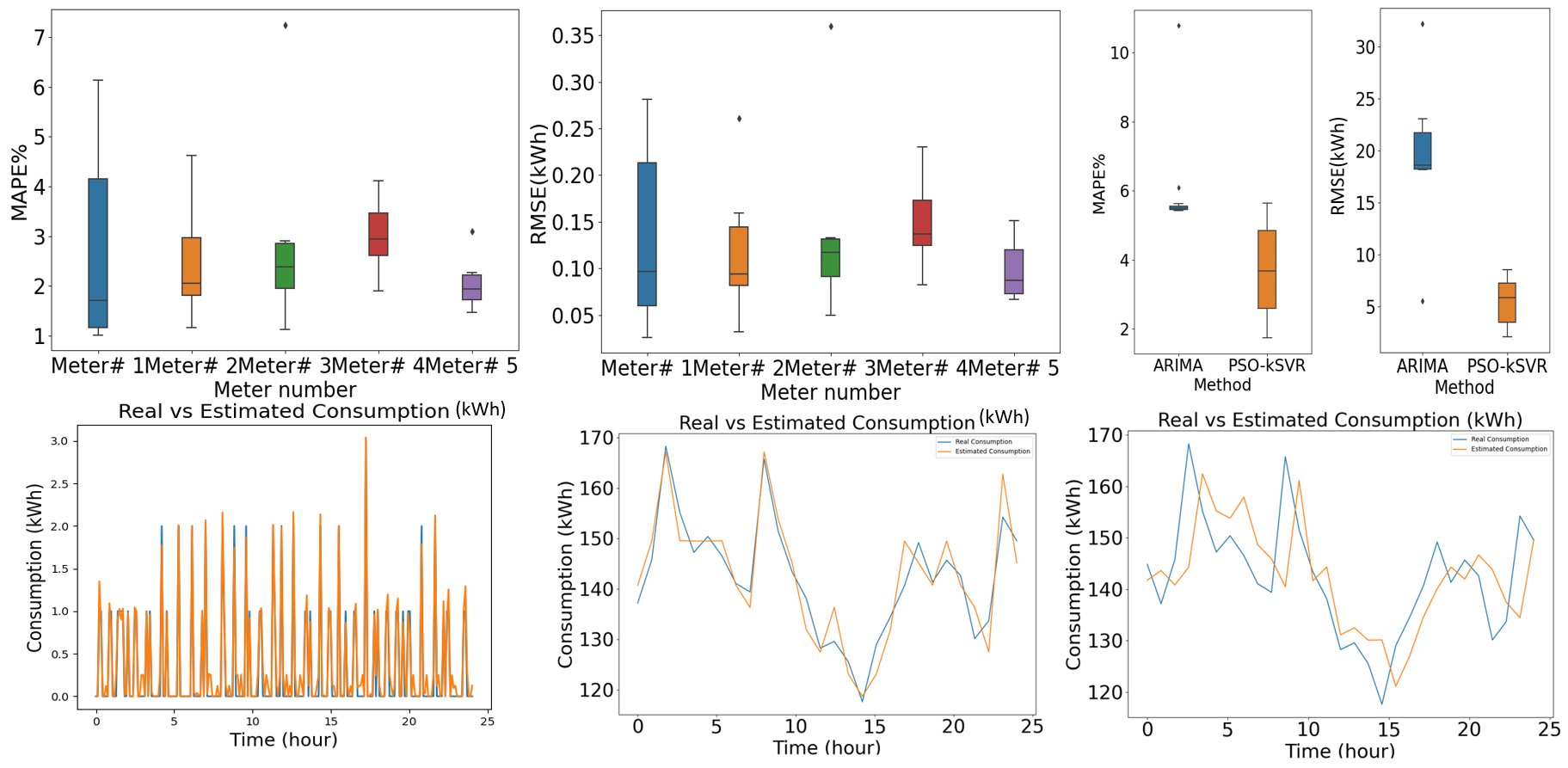}
    \caption{Quantitative and exemplary qualitative results of all experiments conducted are illustrated in the first and second row, respectively. From left to right, 1\textsuperscript{st} row present MAPE and RMSE of PSO-kSVR for \textbf{meter-specific evaluation} and the comparison of \textbf{societal evaluation}. An example of the qualitative result for \textbf{meter-specific evaluation} is depicted in the first image in the 2\textsuperscript{nd} row. The comparison of PSO-kSVR and ARIMA are presented in the 2\textsuperscript{nd} and 3\textsuperscript{rd} images in the lower row. }
    \label{fig:all_results}
 \end{figure}

\textbf{Results} The quantitative and qualitative results both experiments are depicted in Fig.~\ref{fig:all_results}.
For \textbf{meter-specific evaluation}, the range of the mean MAPE is $2.662\pm 0.353\%$ (refer to Fig.\ref{fig:all_results}), whereas the range of RMSE is $0.1288\pm 0.0769$ kWh (refer to Fig.~\ref{fig:all_results}). For the \textbf{societal evaluation}, 40 datasets that have 12 months of historical data are selected and summed up. The quantitative result shows that for the proposed PSO-kSVR the range of MAPE and RMSE is $3.688\pm1.24\%$ and $5.469\pm2.020 $kWh, respectively, whereas the MAPE and RSME of ARIMA are $5.745\pm1.056\%$ and $19.589\pm4.199$ kWh respectively.  

\textbf{Discussion} The results of all experiments conducted indicate that the proposed PSO-kSVR has the capability to perform accurate forecasts for heat load acquired by smart meters and outperforms the state-of-the-art ARIMA model consistently. 
Our PSO-kSVR model exploits the linear trend of the accumulated consumption data, taking the external meteorological factor such as feel-like temperature into consideration. The accumulation of the consumption data, which leads to the reduction of high-frequency noise through low-pass filtering, plays a key role in enhancing the model performance since the expected prediction error decreases if the variance of noise in the underlying data is small \cite{HT}. Additionally, the nonlinear seasonality is modeled with the RBF kernel-based kSVR model, where the optimal hyper-parameters are estimated with PSO. Thus, the proposed methods lead to promising quantitative and qualitative results.
One major limitation of this study is the use of MAPE as an evaluation metrics for quantitative analysis. By definition, MAPE is skewed, i.e.~large errors in the nominator divided by a small denominator represent the ground truth data affect the MAPE dramatically. 
Furthermore, MAPE has an asymmetric tendency of penalization~\cite{bib13} when comparing positive and negative errors, i.e.~over-predictions are punished heavier than under-prediction. Finally, the calculation of MAPE also faces the problem of the division by zero. 
For instance, in the meter-specific evaluation, the ground truth data of each individual smart meter for a certain timestamp is zero, if no heat energy is consumed. In order to tackle this challenge, we add an offset equal to the dynamic range of the respective consumption to the ground truth and predicted value. This consequently introduced bias in the evaluation results. Thus, MAPE is not an ideal evaluation metric to analyse the accuracy of time series prediction methods.

\section{Conclusion and Outlook}
In this work, a novel PSO-kSVR based framework for the prediction of heat consumption at the household level is proposed using accumulated heat consumption data acquired with smart meters as input data. Hereby, the hyper-parameters of the kSVR model are estimated with PSO without any manual parameter tuning process. 
The qualitative and quantitative evaluation results in two different experiment settings demonstrate, that the proposed method is able to achieve accurate prediction of heat consumption for 24 hours with a relatively small training dataset (i.e.~14 days) regardless of data acquisition hardware and season/month. In the future, a comprehensive comparison with deep learning techniques should be performed. Furthermore, more quantitative experiments with heat consumption data from other geographical regions should be conducted. Especially, the time window of 16 days where the input data are linear and stationary may not valid for all geographical regions. This effect needs to be investigated further. Finally, we recommend developing an energy transport model to describe the individual transport latency from the production plant to each household based on a hybrid data set, comprising the asset management data and smart meter data. The combination of this energy transport latency model and the prediction determined based on our proposed method is able to provide utility companies an intuitive data-driven tool towards optimal energy supply with reduced CO\textsubscript{2} emissions.

\medskip
\bibliography{references}

\begin{thebibliography}{18}
\providecommand{\natexlab}[1]{#1}
\providecommand{\url}[1]{\texttt{#1}}
\expandafter\ifx\csname urlstyle\endcsname\relax
  \providecommand{\doi}[1]{doi: #1}\else
  \providecommand{\doi}{doi: \begingroup \urlstyle{rm}\Url}\fi

\bibitem[Program({\natexlab{a}})]{UNEP_1}
United Nation~Environment Program.
\newblock Cities and climate change.
\newblock
  \url{https://www.unep.org/explore-topics/resource-efficiency/what-we-do/cities/cities-and-climate-change},
  {\natexlab{a}}.
\newblock [Online; accessed 08-September-2021].

\bibitem[Program({\natexlab{b}})]{UNEP_2}
United Nation~Environment Program.
\newblock District energy: a secret weapon for climate action and human health.
\newblock
  \url{https://www.unep.org/news-and-stories/story/district-energy-secret-weapon-climate-action-and-human-health},
  {\natexlab{b}}.
\newblock [Online; accessed 08-September-2021].

\bibitem[Bianchi et~al.(2019)Bianchi, Castellini, Tarocco, and Farinelli]{ar}
Federico Bianchi, Alberto Castellini, Pietro Tarocco, and Alessando Farinelli.
\newblock \emph{Load Forecasting in District Heating Networks: Model Comparison
  on a Real-World Case Study}, pages 553--565.
\newblock 01 2019.
\newblock ISBN 978-3-030-37598-0.
\newblock \doi{10.1007/978-3-030-37599-7_46}.

\bibitem[G.E.~Box and Ljung(2015)]{bib4}
G.C.~Reinsel G.E.~Box, G.M.~Jenkins and G.M. Ljung.
\newblock \emph{Time series analysis: forecasting and control}.
\newblock John Wiley and Sons, Boston, 2015.

\bibitem[Fang and Lahdelma(2016)]{arima1}
Tingting Fang and Risto Lahdelma.
\newblock Evaluation of a multiple linear regression model and sarima model in
  forecasting heat demand for district heating system.
\newblock \emph{Applied Energy}, 179:\penalty0 544--552, 2016.
\newblock ISSN 0306-2619.
\newblock \doi{https://doi.org/10.1016/j.apenergy.2016.06.133}.
\newblock URL
  \url{https://www.sciencedirect.com/science/article/pii/S0306261916309217}.

\bibitem[Grosswindhager et~al.(2011)Grosswindhager, Roither-Voigt, and
  Kozek]{bib10}
S~Grosswindhager, Andreas Roither-Voigt, and Martin Kozek.
\newblock Online short-term forecast of system heat load in district heating
  networks.
\newblock \emph{proceedings of the 31st international symposium on forecasting,
  Prag, Czech Republic}, 01 2011.

\bibitem[Benalcazar and Kami{\'{n}}ski(2019)]{bib5}
P~Benalcazar and J~Kami{\'{n}}ski.
\newblock Short-term heat load forecasting in district heating systems using
  artificial neural networks.
\newblock volume 214, page 012023. {IOP} Publishing, jan 2019.
\newblock \doi{10.1088/1755-1315/214/1/012023}.
\newblock URL \url{https://doi.org/10.1088/1755-1315/214/1/012023}.

\bibitem[Jie and Siyuan(2018)]{bib2}
Zhang Jie and Wang Siyuan.
\newblock Thermal load forecasting based on pso-svr.
\newblock pages 2676--2680, 2018.
\newblock \doi{10.1109/CompComm.2018.8780847}.

\bibitem[Xue et~al.(2019)Xue, Pan, Lin, Song, Qi, and Wang]{lstm}
Guixiang Xue, Yu~Pan, Tao Lin, Jiancai Song, Chengying Qi, and Zhipan Wang.
\newblock District heating load prediction algorithm based on feature fusion
  lstm model.
\newblock \emph{Energies}, 12\penalty0 (11), 2019.
\newblock ISSN 1996-1073.
\newblock \doi{10.3390/en12112122}.
\newblock URL \url{https://www.mdpi.com/1996-1073/12/11/2122}.

\bibitem[Vapnik et~al.(1996)Vapnik, Golowich, and Smola]{bib11}
Vladimir Vapnik, Steven~E. Golowich, and Alex Smola.
\newblock Support vector method for function approximation, regression
  estimation, and signal processing.
\newblock In \emph{Advances in Neural Information Processing Systems 9}, pages
  281--287. MIT Press, 1996.

\bibitem[LI~Sun WANG~Chao(2017)]{bib7}
CHENG Tao WANG Yiyuan WANG~Ruiqi LI~Sun WANG~Chao, ZHANG Guilin XU~Zhigen.
\newblock Short-term power load forecasting based on support vector regression.
\newblock \emph{Journal of Shandong University(Engineering Science)},
  47\penalty0 (6):\penalty0 52, 2017.
\newblock \doi{10.6040/j.issn.1672-3961.0.2017.376}.
\newblock URL
  \url{http://gxbwk.njournal.sdu.edu.cn/EN/abstract/article_1681.shtml}.

\bibitem[Idowu et~al.(2014)Idowu, Saguna, Åhlund, and Schelén]{saguna}
Samuel Idowu, Saguna Saguna, Christer Åhlund, and Olov Schelén.
\newblock Forecasting heat load for smart district heating systems: A machine
  learning approach.
\newblock In \emph{2014 IEEE International Conference on Smart Grid
  Communications (SmartGridComm)}, pages 554--559, 2014.
\newblock \doi{10.1109/SmartGridComm.2014.7007705}.

\bibitem[Gao et~al.(2020)Gao, Qi, Xue, Song, Zhang, and Yu]{gao}
Xiaoyu Gao, Chengying Qi, Guixiang Xue, Jiancai Song, Yahui Zhang, and Shi-ang
  Yu.
\newblock Forecasting the heat load of residential buildings with heat metering
  based on ceemdan-svr.
\newblock \emph{Energies}, 13\penalty0 (22), 2020.
\newblock ISSN 1996-1073.
\newblock \doi{10.3390/en13226079}.
\newblock URL \url{https://www.mdpi.com/1996-1073/13/22/6079}.

\bibitem[Dahl et~al.(2018)Dahl, Brun, Kirsebom, and Andresen]{svrcal}
Magnus Dahl, Adam Brun, Oliver~S. Kirsebom, and Gorm~B. Andresen.
\newblock Improving short-term heat load forecasts with calendar and holiday
  data.
\newblock \emph{Energies}, 11\penalty0 (7), 2018.
\newblock ISSN 1996-1073.
\newblock \doi{10.3390/en11071678}.
\newblock URL \url{https://www.mdpi.com/1996-1073/11/7/1678}.

\bibitem[Kennedy(2010)]{bib12}
James Kennedy.
\newblock \emph{Particle Swarm Optimization}, pages 760--766.
\newblock Springer US, Boston, MA, 2010.
\newblock ISBN 978-0-387-30164-8.
\newblock \doi{10.1007/978-0-387-30164-8_630}.
\newblock URL \url{https://doi.org/10.1007/978-0-387-30164-8\_630}.

\bibitem[Hastie et~al.(2009)Hastie, Tibshirani, and Friedman]{HT}
Trevor Hastie, Robert Tibshirani, and Jerome Friedman.
\newblock \emph{Overview of Supervised Learning}, pages 9--41.
\newblock Springer New York, New York, NY, 2009.
\newblock ISBN 978-0-387-84858-7.
\newblock \doi{10.1007/978-0-387-84858-7_2}.
\newblock URL \url{https://doi.org/10.1007/978-0-387-84858-7_2}.

\bibitem[Fan et~al.(2017)Fan, Peng, Zhao, and Hong]{bib1}
Guo-Feng Fan, Li-Ling Peng, Xiangjun Zhao, and Wei-Chiang Hong.
\newblock Applications of hybrid emd with pso and ga for an svr-based load
  forecasting model.
\newblock \emph{Energies}, 10\penalty0 (11), 2017.
\newblock ISSN 1996-1073.
\newblock \doi{10.3390/en10111713}.
\newblock URL \url{https://www.mdpi.com/1996-1073/10/11/1713}.

\bibitem[Goodwin and Lawton(1999)]{bib13}
Paul Goodwin and Richard Lawton.
\newblock On the asymmetry of the symmetric mape.
\newblock \emph{International Journal of Forecasting}, 15\penalty0
  (4):\penalty0 405--408, 1999.
\newblock ISSN 0169-2070.
\newblock \doi{https://doi.org/10.1016/S0169-2070(99)00007-2}.
\newblock URL
  \url{https://www.sciencedirect.com/science/article/pii/S0169207099000072}.

\end{thebibliography}
\clearpage
\appendix

\section{PSO-kSVR Mathematical Modeling}
In general, we use kSVR to perform the forecasting task. Hereby, the heat load prediction is formulated as an optimization task, which minimizes the square of L2-norm of the coefficient vector ($\vec{w}$) of the features such that the error between the target and predicted accumulated consumption is within a margin of error (refer to Eq.~\ref{eq:svr_3}). 
\begin{equation}
    \label{eq:svr_3}
\min   \frac {1}{2}\|\vec{w}\|^2 + C\sum_{i=1}^m (\zeta_{i} + \zeta_{i}^*) ,					
\end{equation}
such that,  		
\begin{equation}
    \label{eq:svr_4}					
    \mid y_i -\vec{w}^T \phi(\vec{x}) - b\mid < \epsilon \ +\zeta_{i}^*.
\end{equation}
\begin{equation}
    \label{eq:svr_5}					
    \mid y_i - \vec{w}^T \phi(\vec{x}) - b\mid < \epsilon \ +\zeta_{i} ,
\end{equation}
\begin{equation}
    \label{eq:svr_6}					
    \zeta_{i}, \zeta_{i}^* < 0 .
\end{equation}
Here, $w_{i}$ are the coefficients or weights of the model and $\epsilon$ is the margin of error that is allowed, known as slack variable. In order to minimize the deviation from the margin, some relaxation factors $\zeta$, $\zeta^*$ are included. $C>0$, the so called box constraint acts as a penalty coefficient that represents the strength of the regulation of observations that lie outside the margin $\epsilon$. 
The training samples are $\lbrace(\vec{x_1}, y_1), (\vec{x_2}, y_2), (\vec{x_3}, y_3)....(\vec{x_m}, y_m)\rbrace, \vec{x_{i}} \in\mathbb{R}^2, y_{i} \in \mathbb{R}$ for $i=1,2,3,...,m$, where $m$ is the number of training samples available. Each feature vector is defined as $\vec{x_{i}}:({h_i}^{t-1}, {\theta_i}^{t-1})$ and the accumulated consumption which we intend to predict is $y_{i}$. Here $\phi(\vec{x})$ is the kernel function $K(\vec{x}, \vec{x}^\prime)$ to transform a non-linear problem into a linear problem by projecting of features into a higher dimension. The kernel employed here is the RBF kernel (also known as Guassian kernel), that is defined as $\exp(-\gamma \|\vec{x} - \vec{x}^\prime \| ^2)$, where $\gamma$ is a hyper-parameter that control the impact of each training example on the overall model. 

Obviously, the choice of the hyper-parameters $\epsilon$, $C$, and $\gamma$ affects the model performance greatly. SVR model with hard-coded hyper-parameters lacks the adaptive capacity to deal with the variance of the input data. Furthermore, manual tuning of the hyper-parameters reduces the usability of the kSVR model for real-world applications. Therefore, a sophisticated adaptive method to determine the optimal configuration of the hyper-parameters for kSVR is incorporated in our method.

Previously, it has been demonstrated that PSO is a promising method to find the optimal model hyper-parameters for high frequent data \cite{bib1}. As the underlying data of our application consists of high-frequent components, we use PSO to optimize the hyper-parameters in a three-dimensional search space, spanned by $C$, $\gamma$, and $\epsilon$. 

On basic level, the PSO model aims to produce an optimal population for the configuration of kSVR by updating the velocity and position of particles as follows:
\begin{equation}
\label{eq:svr_11}
 v_{i}^{j+1} = wv_{i}^j + c_1 r_1 (L_{i}^{j*} - p_{i}^j) + c_2 r_2(G_{i}^{j*} - p_{i}^j),
\end{equation}

\begin{equation}
\label{eq:svr_11}
 p_{i}^{j+1} = p_{i}^j + v_{i}^j .
\end{equation} 
Here, $v_{i}^{j+1}$ and $v_{i}^j$ denotes the new and current velocity, $p_{i}^{j+1}$ and $p_{i}^j$ represents the new and current position respectively for $i^{th}$  particle where $j$ is the iteration number. Also $w$ is the so called inertial weight that defines the search space of the particles, $c_1,c_2$ are the particle's personal and societal learning rate, $r1,r2$ are random numbers uniformly distributed in (0,1) and randomly updated with velocity updates, that represents the weights of two distances updating the particle velocity. Finally, $L_{i}^{j*}, G_{i}^{j*}$ are the current local and global best position, for which minimum prediction error is attained by the model.

In total, we utilize the mean-squared error of predictions as a fitness function for the evaluation of the particles in the population. The algorithm terminates when the iteration limit is reached. At the end of PSO training, the particles converge to such a position in their 3-dimensional search space, i.e.~a combination of $C$, $\gamma$ and $\epsilon$ for which the mean squared error to predict the accumulated consumption using SVR is minimum.

The final heat consumption of each timestamp in the prediction time period, i.e.~heat load, is estimated by computing the first-order derivative of the model output and performing the time shift correction. The latter is necessary because the model has a dependency on historical data to predict future consumption, i.e.~the model always uses the value at the time step $t$ to predict the value at $(t+\tau)^{th}$ instant. This lag can be compensated using time-shifting by the same $\tau$ unit of time.

\section{Experiments Details}
In all experiements conducted, the configuration details of PSO is the following: 20 particles are used in a three-dimensional search space with the inertia of 1 for 50 iterations. The social factors, $c_1$ and $c_2$ are instantiated with 2. The range of $C$, $\epsilon$ and $\gamma$ are initialized as $[1e-3, 1e5]$, $[1e-8, 1e-1]$ and $[1e-3, 1e3]$ respectively. To evaluate the performance of estimation we have used root mean squared error (RMSE), given by the formula,
\begin{equation}
    \label{eq:rmse}
 RMSE =\sqrt{\frac{1}{N}\sum_{i=1}^{N} ( \hat{y_{i}} - y_{i} )^2},		
\end{equation}
where $N$ is the number of samples in prediction window, $\hat{y_{i}}$ is the estimated consumption and $y_{i}$ is the actual consumption. As a comparative evaluation metrics of performance, we have used mean absolute percentage error (MAPE) which is formulated as,
\begin{equation}
    \label{eq:mape}
 MAPE\% = \frac{1}{N}\sum_{i=1}^{N}\frac{\left | \hat{y_{i}} - y_{i} \right |	}{y_{i}} \times 100\%			
\end{equation}

\section{Decomposition of Accumulated Consumption Data}
\begin{figure}[h!]
\centering
\includegraphics[width=12cm]{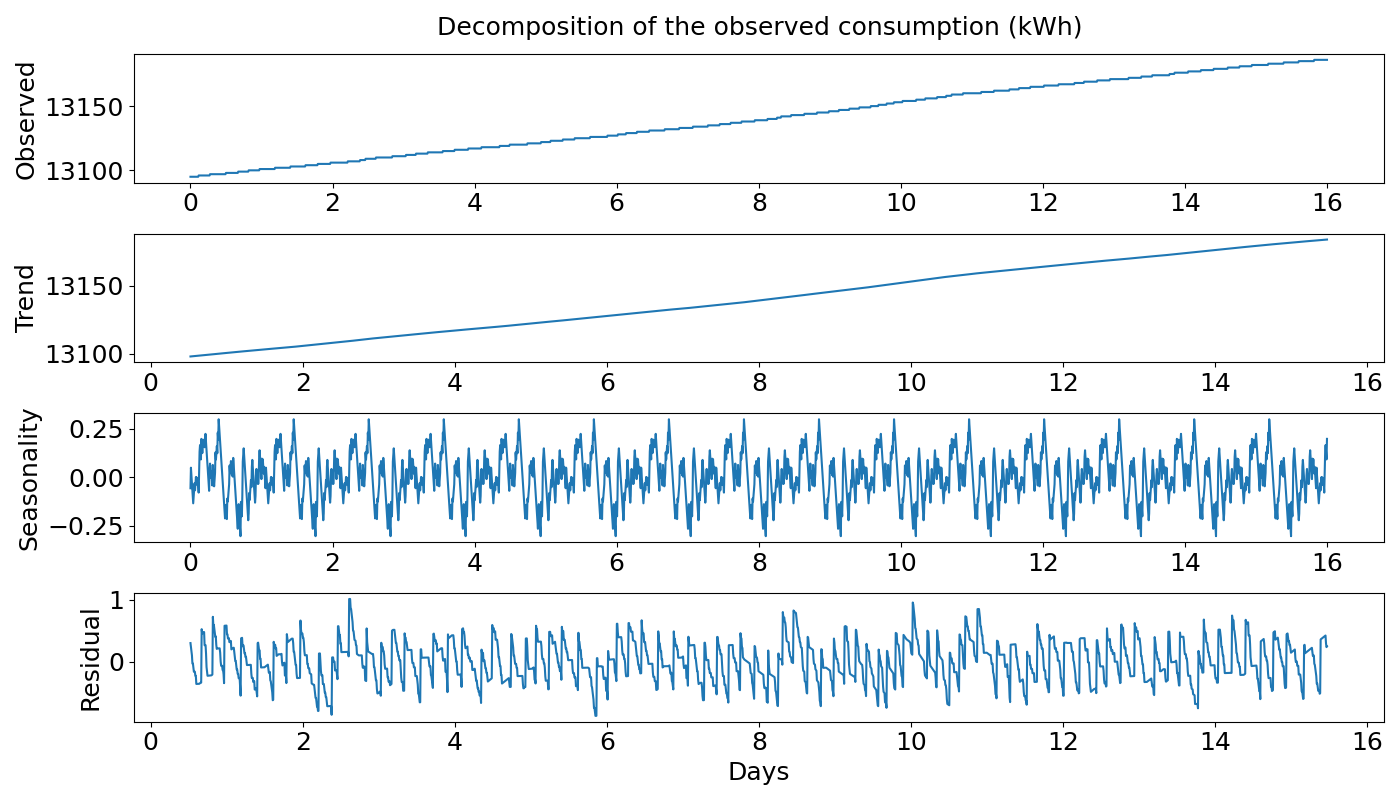}
\caption{Decomposing the time series of consumption into trend, seasonality and residual}
\label{fig:seas_new}
\end{figure}

\section{Further Results}
\begin{figure}[h!]
    \includegraphics[width=14cm]{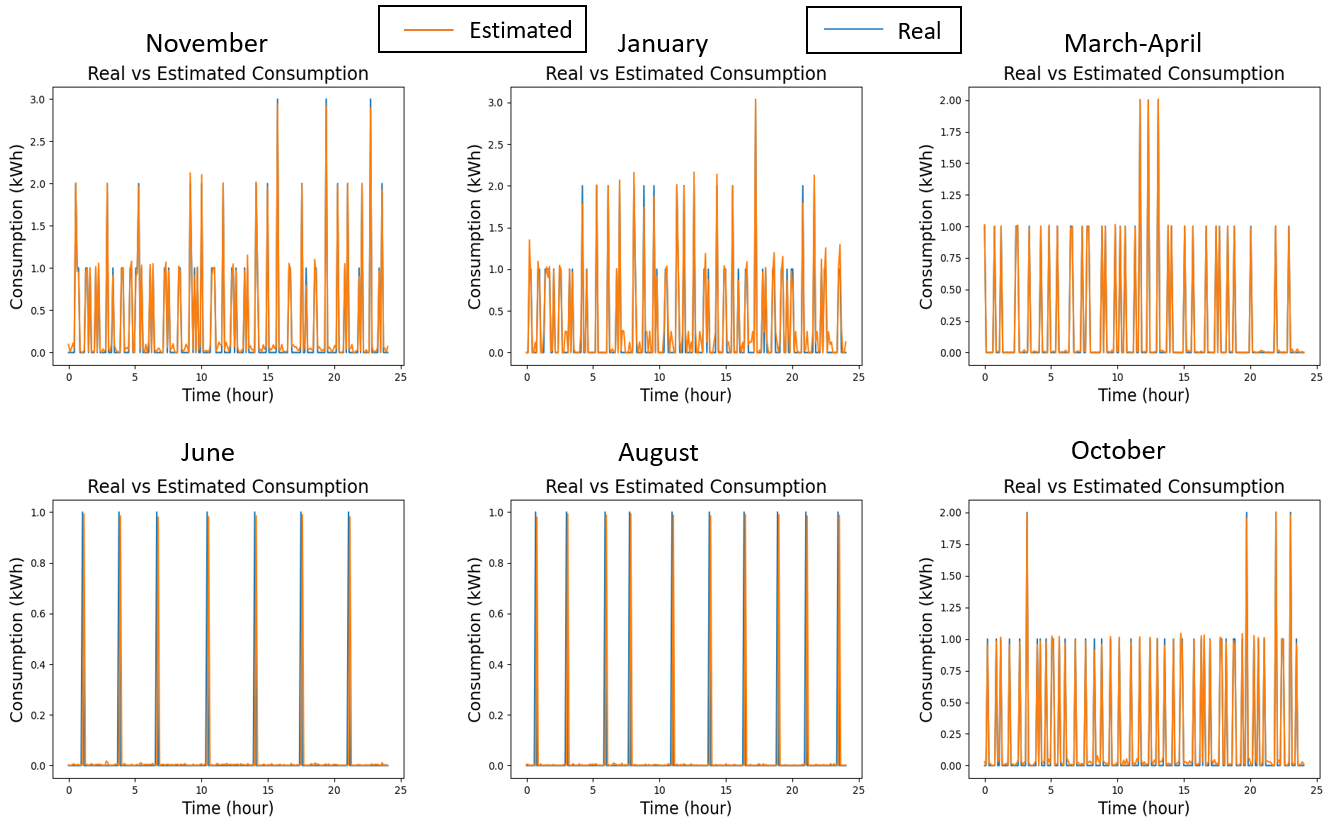}
    \caption{Prediction of heat consumption in different months of the year for an individual meter.}
    \label{fig:monthly_consumption}
 \end{figure}
\begin{figure}[h!]
\centering
\includegraphics[width=14cm]{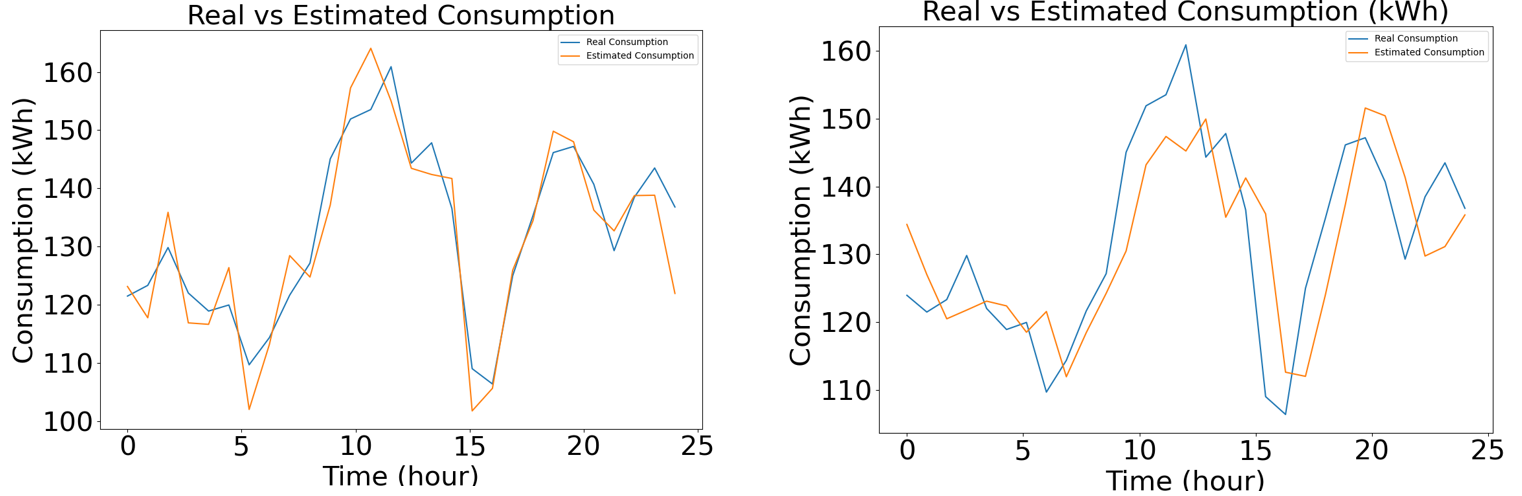}
\caption{Quantitative result of the \textbf{societal experiment}. Forecasting results of PSO-kSVR and ARIMA for the month June are depicted in the left and right images, respectively.}
 \label{fig:comp_jun}
\end{figure}

\begin{table}[h!]
\centering
\caption{Quantitative comparison of ARIMA model and SVR-PSO throughout one year for the \textbf{societal evaluation}.}
  \begin{tabular}{lllll}
    \toprule
    \multirow{2}{*}{Month} &
      \multicolumn{2}{c}{ARIMA model} &
      \multicolumn{2}{c}{SVR-PSO} \\
      & {MAPE\%} & {RMSE} & {MAPE\%} & {RMSE} \\
      \midrule
    November & 8.431$\pm$2.345 & 18.884$\pm$13.344 & \textbf{4.079}$\pm$1.285 & \textbf{2.607}$\pm$0.527 \\
    December & 5.546$\pm$0.017 & 22.047$\pm$0.266 & \textbf{3.388}$\pm$0.682 & \textbf{5.467}$\pm$0.715 \\
    January & 5.530$\pm$0.033 & 19.934$\pm$1.78 & \textbf{3.320}$\pm$1.128 & \textbf{5.381}$\pm$1.955 \\
    February & 5.479$\pm$0.022 & 18.256$\pm$0.002 & \textbf{3.448}$\pm$1.324 & \textbf{5.222}$\pm$1.910 \\
    March & 5.486$\pm$0.028 & 18.848$\pm$0.584 & \textbf{4.084}$\pm$0.912 & \textbf{6.268}$\pm$1.846 \\
    April & 5.549$\pm$0.021 & 18.243$\pm$0.022 & \textbf{3.707}$\pm$1.206 & \textbf{5.716}$\pm$1.812 \\
    May & 5.473$\pm$0.029 & 19.354$\pm$1.121 & \textbf{5.114}$\pm$0.281 & \textbf{7.764}$\pm$0.545 \\
    June & 5.540$\pm$0.007 & 18.500$\pm$0.22 & \textbf{2.920}$\pm$0.293 & \textbf{4.890}$\pm$0.67 \\
    July & 5.460$\pm$0.028 & 19.566$\pm$0.145 & \textbf{2.636}$\pm$0.899 & \textbf{4.434}$\pm$2.275 \\
    August & 5.518$\pm$0.049 & 20.637$\pm$2.327 & \textbf{3.951}$\pm$1.696 & \textbf{6.017}$\pm$2.547 \\
    September & 5.441$\pm$0.002 & 20.642$\pm$2.41 & \textbf{4.377}$\pm$0.546 & \textbf{6.852}$\pm$0.636 \\
    October & 5.587$\pm$0.042 & 20.158$\pm$1.692 & \textbf{3.231}$\pm$1.348 & \textbf{5.016}$\pm$2.092 \\
    \bottomrule
  \end{tabular}
  \label{fig:table_perf}
\end{table}

\end{document}